\documentclass[runningheads]{llncs}

\usepackage{tikz}
\usepackage{comment}
\usepackage{color}
\usepackage{graphicx}
\usepackage{epsfig}
\usepackage{amsmath}
\usepackage{amssymb}
\usepackage{booktabs}
\usepackage{physics}
\usepackage{mathtools}
\usepackage[ruled, noend]{algorithm2e}
\usepackage{mwe}
\usepackage{caption}
\usepackage{multirow}
\usepackage{makecell}
\usepackage{hhline}
\usepackage{diagbox}
\usepackage{adjustbox}
\usepackage[accsupp]{axessibility}  

\renewcommand{\vec}[1]{\boldsymbol{#1}}

\DeclarePairedDelimiterX{\infdivx}[2]{(}{)}{%
  #1\;\delimsize\|\;#2%
}

\SetKwIF{If}{ElseIf}{Else}{if}{}{else if}{else}{end if}

\makeatletter
\DeclareRobustCommand\onedot{\futurelet\@let@token\@onedot}
\def\@onedot{\ifx\@let@token.\else.\null\fi\xspace}

\def\eg{\emph{e.g}\onedot}

\def\etc{\emph{etc}\onedot} 
\def\wrt{w.r.t\onedot} 
 
\def\etal{\emph{et al}\onedot}
\makeatother

\begin{document}
\pagestyle{headings}
\mainmatter
\def\ECCVSubNumber{1991}  

\title{TOCH: Spatio-Temporal Object-to-Hand Correspondence for Motion Refinement} 

\titlerunning{Spatio-Temporal Object-to-Hand Correspondence for Motion Refinement
}
%
\author{Keyang Zhou\inst{1,2} \and
Bharat Lal Bhatnagar \inst{1,2} \and \\
Jan Eric Lenssen\inst{2} \and Gerard Pons-Moll\inst{1,2}}
\authorrunning{Zhou et al.}
%
\institute{University of Tübingen, Germany
\email{\{keyang.zhou,gerard.pons-moll\}@uni-tuebingen.de}
\and
 Max Planck Institute for Informatics, Saarland Informatics Campus, Germany
 \email{\{bbhatnag,jlenssen\}@mpi-inf.mpg.de}
 }

\maketitle

\begin{abstract}
   We present TOCH, a method for refining incorrect 3D hand-object interaction sequences using a correspondence based prior learnt directly from data.
   Existing hand trackers, especially those that rely on very few cameras, often produce visually unrealistic results with hand-object intersection or missing contacts. Although correcting such errors requires reasoning about temporal aspects of interaction, most previous works focus on static grasps and contacts.
   The core of our method are TOCH fields, a novel spatio-temporal representation for modeling correspondences between hands and objects during interaction. 
   TOCH fields are a point-wise, object-centric representation, which encode the hand position relative to the object. Leveraging this novel representation, we learn a latent manifold of plausible TOCH fields with a temporal denoising auto-encoder. 
Experiments demonstrate that TOCH outperforms state-of-the-art 3D hand-object interaction models, which are limited to static grasps and contacts. More importantly, our method produces smooth interactions even before and after contact. 
   Using a single trained TOCH model, we quantitatively and qualitatively demonstrate its usefulness for correcting erroneous sequences from off-the-shelf RGB/RGB-D hand-object reconstruction methods and transferring grasps across objects.
   Our code and model are available at~\cite{toch}.
   
\end{abstract}

\section{Introduction}
\label{sec:intro}
Tracking hands that are in interaction with objects is an important part of many applications in Virtual and Augmented Reality, such as modeling digital humans capable of manipulation tasks~\cite{starke2019neural,guzov22hops,zhang2022couch,bhatnagar22behave,yi2022mover}. Although there exists a vast amount of literature about tracking hands in isolation, much less work has focused on joint tracking of objects and hands. The high degrees of freedom in possible hand configurations, frequent occlusions, noisy or incomplete observations (\emph{e.g} lack of depth channel in RGB images) make the problem heavily ill-posed.
We argue that tracking interacting hands requires a powerful prior learned from a set of clean interaction sequences, which is the core principle of our method.

\begin{figure*}[t]
    \centering
    \includegraphics[width=\linewidth]{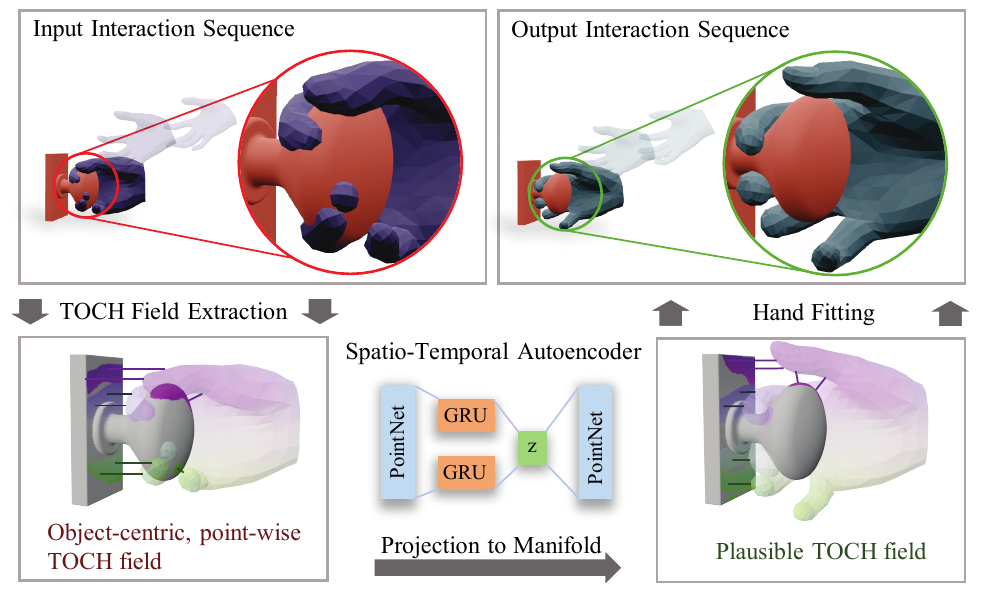}
    \caption{We propose TOCH, a model for correcting erroneous hand-object interaction sequences. TOCH takes as input a tracking sequence produced by any existing tracker.
    We extract TOCH fields, a novel object-centric correspondence representation, from a noisy hand-object mesh sequence.
    The extracted noisy TOCH fields are fed into an auto-encoder, which projects it onto a learned hand motion manifold. Lastly, we obtain the corrected tracking sequence by fitting hands to the reconstructed TOCH fields. TOCH is applicable to interaction sequences even before and after contact happens. }
    \label{fig:overview}
\end{figure*}

\begin{figure*}[t]
    \centering
    \includegraphics[width=\linewidth]{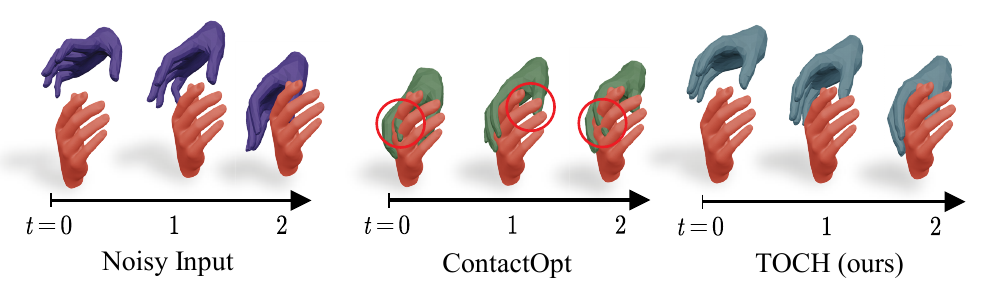}
    \caption{Example refinement of an interaction sequence. Left: a noisy sequence of a hand approaching and grasping another static hand. Middle: ContactOpt~\cite{grady2021contactopt} always snaps the hand into grasping posture regardless of its position relative to the object, as it is not designed for sequences. Right: TOCH preserves the relative hand-object arrangement during interaction while refining the final grasp.}
    \label{fig:snapping}
\end{figure*}

Beyond the aforementioned challenges, subtle errors in hand estimation have a huge impact on perceived realism. For example, if the 3D object is floating in the air, is grasped in a non-physically plausible way, or hand and object intersect, the perceived quality will be poor. Unfortunately, such artifacts are common in pure hand-tracking methods. Researchers have used different heuristics to improve plausibility, such as inter-penetration constraints~\cite{hasson19obman} and smoothness priors~\cite{hasson2020leveraging}. A recent line of work predicts likely static hand poses and grasps for a given object~\cite{karunratanakul2020grasping,hamer2010object} but those methods can not directly be used as a prior to fix common capturing and tracking errors.
Although there exists work to refine hand-object interactions~\cite{taheri2020grab,grady2021contactopt}, it is only concerned with static grasps.

In this work, we propose TOCH, a data-driven method for refining noisy 3D hand-object interaction sequences. In contrast to previous work in interaction modeling, TOCH not only considers static interactions but can also be applied to sequences without introducing snapping artifacts.
The whole approach is outlined in Figure~\ref{fig:overview}. Our key insight is that estimating point-wise, spatio-temporal object-hand correspondences are crucial for realism, and sufficient to constrain the high-dimensional hand poses. 
Thus, the point-wise corresponcendes between object and hand are encoded in a novel spatio-temporal representation called \emph{TOCH field}, which takes the object geometry and the configuration of the hand with respect to the object into account. We then learn the manifold of plausible TOCH fields from the recently released MoCap dataset of hand-object interactions~\cite{taheri2020grab} using an auto-encoder and apply it to correcting noisy observations.
In contrast to conventional binary contacts~\cite{brahmbhatt2020contactpose,grady2021contactopt,xie22chore}, TOCH fields also encode the position of hand parts that are not directly in contact with the object, making TOCH applicable to whole interaction sequences, see Figure~\ref{fig:snapping}.
TOCH has further useful properties for practical application:
\begin{itemize}
    \item TOCH can effectively project implausible hand motions to the learned object-centric hand motion manifold and produces visually correct interaction sequences that outperform previous static approaches.

    \item TOCH does not depend on specific sensor data (RGB image, depth map \etc) and can be used with any tracker.
    \item TOCH can be used to transfer grasp sequences across objects sharing similar geometry, even though it is not designed for this task.
\end{itemize}

\section{Related Work}
\label{sec:related}
\subsection{Hand and Object Reconstruction}
\label{sec:related_recon}
\subsubsection{Hand Reconstruction and Tracking.}
Reconstructing 3D hand surfaces from RGB or depth observations is a well-studied problem~\cite{huang2021survey}. Existing work can generally be classified into two paradigms: discriminative approaches~\cite{ge20193d,chen2021mvhm,zhao2020hand,malik2020handvoxnet,boukhayma20193d} directly estimate hand shape and pose parameters from the observation, while generative approaches~\cite{sridhar2014real,taylor2016efficient,taylor2017articulated} iteratively optimize a parametric hand model so that its projection matches the observation. Recently, more challenging settings such as reconstructing two interacting hands~\cite{zhang2021interacting,mueller2019real,smith2020constraining} are also explored. These works ignore the presence of objects and are hence less reliable in interaction-intensive scenarios.

\vspace{0.2cm}
\noindent\textbf{Joint Hand and Object Reconstruction.}
\label{sec:related_jointrecon}
Jointly reconstructing hand and object in interaction~\cite{oikonomidis2011full,ballan2012motion,wang2013video,sridhar2016real,zhang2019interactionfusion,zhang2021single,panteleris2017back} has received much attention. Owing to the increasing amount of hand-object interaction datasets with annotations~\cite{hasson2019learning,hampali2020honnotate,brahmbhatt2020contactpose,garcia2018first,zimmermann2019freihand,kwon2021h2o}, deep neural networks are often used to estimate an initial hypothesis for hand and object poses, which are then jointly optimized to meet certain interaction constraints~\cite{hasson2019learning,hasson2020leveraging,chen2021joint,hasson2021towards,cao2021reconstructing}. Most works in this direction improve contact realism by encouraging a small hand-to-object distance and penalizing inter-penetrating vertices~\cite{jiang2020coherent,ballan2012motion}. However, these simple approaches often yield implausible interaction and do not take the whole motion sequence into account. In contrast, our method alleviates both shortcomings through a object-centric, temporal representation that also considers frames in which hand and object are not in direct contact.

\subsection{Hand Contact and Grasp}
\label{sec:related_grasp}

\noindent\textbf{Grasp Synthesis.}
Synthesizing novel hand grasp given an object has been widely studied in robotics~\cite{sahbani2012overview}. Traditional approaches either optimize for force-closure condition~\cite{el20113d} or sample and rank grasp candidates based on learned features~\cite{bohg2013data}. There are also hybrid approaches that combine the merits of both~\cite{miller2004graspit,leon2010opengrasp}. Recently, a number of neural network-based models have been proposed for this task~\cite{hasson19obman,taheri2020grab,corona2020ganhand,zhu2021toward,jiang2021hand}. In particular, \cite{karunratanakul2020grasping,jiang2021synergies} represent the hand-object proximity as an implicit function. We took a similar approach and represent the hand relative to the object by signed distance values distributed on the object.

\vspace{0.2cm}
\noindent\textbf{Object Manipulation Synthesis.}
In comparison with static grasp synthesis, generating dexterous manipulation of objects is a more difficult problem since it additionally requires dynamic hand and object interaction to be modeled. This task is usually approached by optimizing hand poses to satisfy a range of contact force constraints~\cite{liu2009dextrous,ye2012synthesis,mordatch2012contact,zhao2013robust}. Hand motions generated by these works are physically plausible but lack natural variations. Zhang~\etal~\cite{zhang2021manipnet} utilized various hand-object spatial representations to learn object manipulation from data. An IK solver is used to avoid inter-penetration. We took a different approach and solely use an object-centric spatio-temporal representation, which is shown to be less prone to interaction artifacts.

\vspace{0.2cm}
\noindent\textbf{Contact Refinement.}
Recently, some works focus on refining hand and object contact~\cite{taheri2020grab,yang2021cpf,grady2021contactopt}. Both~\cite{yang2021cpf} and~\cite{grady2021contactopt} propose to first estimate the potential contact region on the object and then fit the hand to match the predicted contact. However, limited by the proposed contact representation, they can only model hand and object \emph{in stable grasp}. While we share a similar goal, our work can also deal with the case where the hand is \emph{close to} but not in contact with the object, as a result of our novel hand-object correspondence representation. Hence our method can be used to refine a tracking sequence.

\subsection{Pose and Motion Prior}
\label{sec:related_prior}
It has been observed that most human activities lie on low-dimensional manifolds~\cite{elgammal2008role,urtasun20063d}. Therefore natural motion patterns can be found by applying learned data priors. A pose or motion prior can facilitate a range of tasks including pose estimation from images or videos~\cite{bogo2016keep,arnab2019exploiting,luo20203d,zeng2022smoothnet}, motion interpolation~\cite{li2021task}, motion capture~\cite{zhang2021learning,tiwari22posendf}, and motion synthesis~\cite{henter2020moglow,aliakbarian2020stochastic,cai2021unified}. Early attempts in capturing pose and motion priors mostly use simple statistical models such as PCA~\cite{ormoneit2001learning}, Gaussian Mixture Models~\cite{bogo2016keep} or Gaussian Process Dynamical Models~\cite{urtasun20063d}. With the advent of deep generative models~\cite{kingma2013auto,goodfellow2014generative}, recent works rely on auto-encoders~\cite{pavlakos2019expressive,kocabas2020vibe} and adversarial discriminators~\cite{zhao2020bayesian,kundu2019bihmp} to more faithfully capture the motion distribution.

Compared to body motion prior, there is less work devoted to hand motion priors. Ng~\etal\cite{ng2021body2hands} learned a prior of conversational hand gestures conditioned on body motion. Our work bears the most similarity to~\cite{hamer2010object}, where an object-dependent hand pose prior was learned to foster tracking. Hamer~\etal~\cite{hamer2010object} proposed to map hand parts into local object coordinates and learn the object-dependent distribution with a Parzen density estimator. The prior is learned on a few objects and subsequently transferred to objects from the same class by geometric warping. Hence it cannot truly capture the complex correlation between hand gesture and object geometry.

\section{Method}
\label{sec:method}
In this section, we describe our method for refining hand pose sequences during interaction with an object. We begin by introducing the problem setting and outlining our approach.
Let $\vec{H}= (\vec{H}^i)_{1\leq i \leq T}$ with $\vec{H}^i \in\mathbb{R}^{K\times 3}$ denote a sequence of vertices that describe hand meshes over the course of an interaction over $T$ frames. We only deal with sequences containing a single hand and a single rigid object mesh, whose vertices we denote as $\vec{O}\in\mathbb{R}^{L\times 3}$. We assume the object shape to be known. Since we care about hand motion relative to the object, we express the hands in local object space, and the object coordinates remain fixed over the sequence. The per-frame hand vertices $\vec{H}^i$ in object space are produced by a parametric hand model MANO~\cite{MANO:SIGGRAPHASIA:2017} using linear blend skinning:
\begin{equation}
\label{eq:mano}
    \vec{H}^i = \text{LBS}\left(\vec{Y};\vec{\beta} ,\vec{\theta}^i \right) + \vec{t}_H^i.
\end{equation}
where the parameters $\{ \vec{\beta}^i, \vec{\theta}^i, \vec{t}^i \}$ denote shape, pose and translation \wrt template hand mesh $\vec{Y}$ respectively.

Observing the hand-object motion through RGB or depth sensors, a hand tracker yields an estimated hand motion sequence $\tilde{\vec{H}}= (\tilde{\vec{H}}^i)_{1\leq i \leq T}$. The goal of our method is to improve the perceptual realism of this potentially noisy estimate using prior information learned from training data.

\vspace{0.2cm}
\noindent
\textbf{Concept.}
We observe that during hand-object interactions, the hand motion is heavily constrained by the object shape. Therefore, noisy hand-object interaction is a deviation from a low-dimensional manifold of realistic hand motions, conditioned on the object. We formulate our goal as learning a mapping to maximize the posterior $p(\vec{H} | \tilde{\vec{H}}, \vec{O})$ of the real motion $\vec{H}$ given the noisy observation $\tilde{\vec{H}}$ and the object with which the hand interacts. This amounts to finding an appropriate sequence of MANO parameters, which is done in three steps (see Figure~\ref{fig:overview}): \textbf{1)} The initial estimate of a hand motion sequence is encoded with a TOCH field, our object-centric, point-wise correspondence representation (Section~\ref{sec:method_repr}). \textbf{2)} The TOCH fields are projected to a learned low-dimensional manifold using a temporal denoising auto-encoder (Section~\ref{sec:method_ae}). \textbf{3)} A sequence of corrected hand meshes is obtained from the processed TOCH fields (Section~\ref{sec:method_recon}).

\subsection{TOCH Fields}
\label{sec:method_repr}
Naively training an auto-encoder on hand meshes is problematic, because the model could ignore the conditioning object and learn a plain hand motion prior (Sec.~\ref{sec:exp_ablation}). Moreover, if we include the object into the formulation, the model would need to learn manifolds for all joint rigid transformation of hand and object, which leads to high problem complexity~\cite{karunratanakul2020grasping}.
Thus, we represent the hand as a TOCH field $\vec{F}$, which is a spatio-temporal object-centric representation that makes our approach invariant to joint hand and object rotation and translation.

\vspace{0.2cm}
\noindent
\textbf{TOCH Field Representation.} For an initial estimation $\tilde{\vec{H}}$ of the hand mesh and the given object mesh $\vec{O}$, we define the TOCH field as a collection of point-wise vectors on a set $\{\vec{o}_i\}_{i=1}^N$ of $N$ points, sampled from the object surface:
\begin{align}
    \vec{F}(\tilde{\vec{H}}, \vec{O}) = \{(c_i, d_i, \vec{y_i})\}_{i=1}^N,
\end{align}
where $c_i \in \{0, 1\}$ is a binary flag indicating whether the $i$-th sampled object point has a corresponding point on the hand surface, $d_i \in \mathbb{R}$ is the signed distance between the object point and its corresponding hand point, and $\vec{y}_i \in \mathbb{R}^3$ are the coordinates of the corresponding hand point on the un-posed canonical MANO template mesh. Note that  $\vec{y}_i$ is a 3D location on the hand surface embedded in $\mathbb{R}^3$, encoding the correspondence similar to~\cite{bhatnagar2020loopreg,taylor2012vitruvian}.

\vspace{0.2cm}
\noindent
\textbf{Finding correspondences.} As we model whole interaction sequences, including frames in which the hand and the object are not in contact, we cannot simply define the correspondences as points that lie within a certain distance to each other. Instead, we generalize the notion of contact by diffusing the object mesh into $\mathbb{R}^3$. We cast rays from the object surface along its normal directions, as outlined in Figure~\ref{fig:overview}. The object normal vectors are obtained from the given object mesh. 
The correspondence of an object point is obtained as the first intersection with the hand mesh. If there is no intersection, or the first intersection is not the hand, this object point has no correspondence.
If the object point is inside the hand, which might happen in case of noisy observations, we search for correspondences along the negative normal direction.
The detailed procedure for determining correspondences is listed in Algorithm~\ref{alg:corr}.

\vspace{0.2cm}
\noindent
\textbf{Representation properties.} The described TOCH field representation has the following advantages. \textbf{1)} It is naturally invariant to joint rotation and translation of object and hand, which reduces required model complexity. 
\textbf{2)} By specifying the distance between corresponding points, TOCH fields enable a subsequent auto-encoder to reason about point-wise proximity of hand and object. This helps to correct various artifacts, \eg inter-penetration can be simply detected by finding object vertices with a negative correspondence distance. \textbf{3)} From surface normal directions of object points and the corresponding distances, a TOCH field can be seen as an encoding of the partial hand point cloud from the perspective of the object surface. We can explicitly derive that point cloud from the TOCH field and use it to infer hand pose and shape by fitting the hand model to the point cloud (c.f. Section~\ref{sec:method_recon}).

\begin{algorithm}[t]
\caption{Finding object-hand correspondences}\label{alg:corr}
\KwIn{Hand mesh $\vec{H}$, object mesh $\vec{O}$, uniformly sampled object points and normals $\{\vec{o}_i, \vec{n}_i\}_{i=1}^N\ $}
\KwOut{Binary correspondence indicators $\{c_i\}_{i=1}^N$}
\For{$i=1$ \KwTo $N$}{
$c_i \gets 0$\;
  \lIf{$\vec{o}_i$ \textup{inside} $\vec{H}$}{
    $s \gets -1$ \textup{\textbf{else}} $s \gets 1$
  }
  $\vec{r}_1 \gets \textup{ray}(\vec{o}_i, s\vec{n}_i)$\;
  $\vec{p}_1 \gets \textup{ray\_mesh\_intersection}(\vec{r}_1, \vec{H})$\;
  \If{$\vec{p}_1 \neq \varnothing$} {
    $\vec{r}_2 \gets \textup{ray}(\vec{o}_i+\epsilon s\vec{n}_i, s\vec{n}_i)$\;
   $\vec{p}_2 \gets \textup{ray\_mesh\_intersection}(\vec{r}_2, \vec{O})$\;
    \If{$\vec{p}_2 = \varnothing$ \textup{\textbf{or}} $\left\| \vec{o}_i - \vec{p}_1 \right\| < \left\|\vec{o}_i - \vec{p}_2\right\|$} {
        $c_i \gets 1$;
    }
  }
}
\end{algorithm}

\subsection{Temporal Denoising Auto-encoder}
\label{sec:method_ae}
To project a noisy TOCH field to the correct manifold, we use a temporal denoising auto-encoder, consisting of an encoder $g_\text{enc}: (\tilde{\vec{F}}_i)_{1\leq i \leq T}\mapsto (\vec{z}_i)_{1\leq i \leq T}$, which maps a sequence of noisy TOCH fields (concatenated with the coordinates and normals of each object point) to latent representation, and a decoder $g_\text{dec}: (\vec{z}_i)_{1\leq i \leq T} \mapsto (\hat{\vec{F}}_i)_{1\leq i \leq T}$, which computes the corrected TOCH fields $\hat{\vec{F}}$ from the latent codes. As TOCH fields consist of feature vectors attached to points, we use a PointNet-like~\cite{qi2017pointnet} architecture. 
The point features in each frame are first processed by consecutive PointNet blocks to extract frame-wise features. These features are then fed into a bidirectional GRU layer to capture temporal motion patterns. The decoder network again concatenates the encoded frame-wise features with coordinates and normals of the object points and produces denoised TOCH fields $(\hat{\vec{F}}_i)_{1\leq i \leq T}$. 
The network is trained by minimizing
\begin{equation}
    \mathcal{L} (\hat{\vec{F}}, \vec{F}) = 
    \sum_{i=1}^T\sum_{j=1}^{N}c_j^i \left(\| \hat{\vec{y}}_j^i - \vec{y}_j^i\|_2^2 + w_{ij}( \hat{d}_j^i - d_j^i)^2 \right) - \text{BCE}(\hat{c}_j^i, c_j^i)  \textnormal{,}
\end{equation}
where $\vec{F}$ denotes the groundtruth TOCH fields and $\text{BCE}(\hat{c}_j^i, c_j^i)$ is the binary cross entropy between output and target correspondence indicators. Note that we only compute the first two parts of the loss on TOCH field elements with $c_j^i = 1$, i.e. object points that have a corresponding hand point.
We use a weighted loss on the distances $\hat{d}_j^i$. The weights are defined as
\begin{align}
    w_{ij} = \frac{\exp \left( -\left\| d_{j}^{i} \right\| \right)}{\sum_{k=1}^{N_i}{\exp \left( -\left\| d_{k}^{i} \right\| \right)}}  N_i \textnormal{,}
\end{align}
where $N_i = \sum_{j=1}^N c_j^i$. This weighting scheme encourages the network to focus on regions of close interaction, where a slight error could have huge impact on contact realism.
Multiplying by the sum of correspondence ensures equal influence of all points in the sequence instead of equal influence of all frames.

\subsection{Hand Motion Reconstruction}
\label{sec:method_recon}
After projecting the noisy TOCH fields of input tracking sequence to the manifold learned by the auto-encoder, we need to recover the hand motion from the processed TOCH fields.
The TOCH field is not fully differentiable w.r.t. the hand parameters, as changing correspondences would involve discontinuous function steps. Thus, we cannot directly optimize the hand pose parameters to produce the target TOCH field.
Instead, we decompose the optimization into two steps. We first use the denoised TOCH fields to locate hand points corresponding to the object points. We then optimize the MANO model to find hands that best fit these points, which is a differentiable formulation.

Formally, given denoised TOCH fields $\vec{F}^i(\vec{H}, \vec{O}) = \{(c_j^i, d_j^i, \vec{y}_j^i)\}_{j=1}^N$ for frames $i\in \{1,...,T\}$ on object points $\{\vec{o}_j\}_{j=1}^N$, we first produce the partial point clouds $\hat{\vec{Y}}^i$ of the hand as seen from the object's perspective:
\begin{equation}
    \hat{\vec{y}}^i_j = \vec{o}_j+d^i_j\vec{n}^i_j \textnormal{.}
\end{equation}
Then, we fit MANO to those partial point clouds by minimizing:
\begin{align}
    \mathcal{L}(\vec{\beta}, \vec{\theta}, \vec{t}_H) = \sum_{i=1}^T \mathcal{L}_\text{corr}(\vec{\beta}, \vec{\theta}^i, \vec{t}_H) + \mathcal{L}_\text{reg}(\vec{\beta}, \vec{\theta}) \textnormal{.}
    \label{eq:loss}
\end{align}
The first term of Equation~\ref{eq:loss} is the hand-object correspondence loss
\begin{equation}
     \mathcal{L}_\text{corr}(\vec{\beta}, \vec{\theta}^i, \vec{t}_H) = \sum_{j=1}^N{c_j^i\left\| \hat{\vec{y}}^i_j - \left(\text{LBS}\left( \textnormal{Proj}_{\vec{Y}} {\left(\vec{y}^i_j\right)};\vec{\beta} ,\vec{\theta}^i \right) + \vec{t}_H\right) \right\| ^2} \textnormal{,}  \label{eq:loss_corr}
\end{equation}
where LBS is the linear blend skinning function in Equation~\ref{eq:mano} and $\textnormal{Proj}_{\vec{Y}}(\cdot)$ projects a point to the nearest point on the template hand surface.
This loss term ensures that the deformed template hand point corresponding to $\vec{o}_i$ is at a predetermined position derived from the TOCH field.

The last term of (\ref{eq:loss}) regularizes shape and pose parameters of MANO,
\begin{equation}
     \mathcal{L}_\text{reg}(\vec{\beta}, \vec{\theta}) = w_1\left\| \vec{\beta} \right\| ^2+w_2\sum_{i=1}^T{\left\| \vec{\theta}^i \right\| ^2} + w_3\sum_{i=1}^{T-1}{\left\| \vec{\theta}^{i+1} - \vec{\theta}^{i} \right\| ^2} + w_4\sum_{i=2}^{T-1}{\sum_{k=1}^J{\left\| \ddot{\vec{p}}_k^i\right\|} }
\end{equation}
where $\ddot{\vec{p}}_k^i$ is the acceleration of hand joint $k$ in frame $i$.
Besides regularizing the norm of MANO parameters, we additionally enforce temporal smoothness of hand poses. This is necessary because (\ref{eq:loss_corr}) only constrains those parts of a hand with object correspondences. Per-frame fitting of TOCH fields leads to multiple plausible solutions, which can only be disambiguated by considering neighbouring frames.
Since (\ref{eq:loss}) is highly nonconvex, we optimize it in two stages. In the first stage, we freeze shape and pose, and only optimize hand orientation and translation. We then jointly optimize all the variables in the second stage.

\section{Experiments}
\label{sec:exp}

In this section, we evaluate the presented method on synthetic and real datasets of hand/object interaction. Our goal is to verify that TOCH produces \emph{realistic interaction sequences} (Section~\ref{sec:exp_synthetic}), \emph{outperforms previous static approaches} in several metrics (Section~\ref{sec:exp_real}), and derives a \emph{meaningful representation} for hand object interaction (Section~\ref{sec:exp_ablation}). Before presenting the results, we introduce the used datasets in Section~\ref{sec:exp_datasets} and the evaluated metrics in Section~\ref{sec:exp_metrics}. 

\subsection{Datasets}
\label{sec:exp_datasets}
\noindent
\textbf{GRAB}. We train TOCH on GRAB~\cite{taheri2020grab}, a MoCap dataset for whole-body grasping of objects. GRAB contains interaction sequences with $51$ objects from~\cite{brahmbhatt2019contactdb}. We pre-select $10$ objects for validation and testing, and train with the rest sequences. Since we are only interested in frames where interaction is about to take place, we filter out frames where the hand wrist is more than $15$ cm away from the object. Due to symmetry of the two hands, we anchor correspondences to the right hand and flip left hands to increase the amount of training data.

\vspace{0.1cm}
\noindent
\textbf{HO-3D}. HO-3D is a dataset of hand-object video sequences captured by RGB-D cameras. It provides frame-wise annotations for 3D hand poses and $6$D object poses, which are obtained from a novel joint optimization procedure. To ensure fair comparison with baselines which are not designed for sequences without contact, we compare on a selected subset of static frames with hand-object contact.

\begin{figure*}[t]
    \centering
    \includegraphics[width=\linewidth]{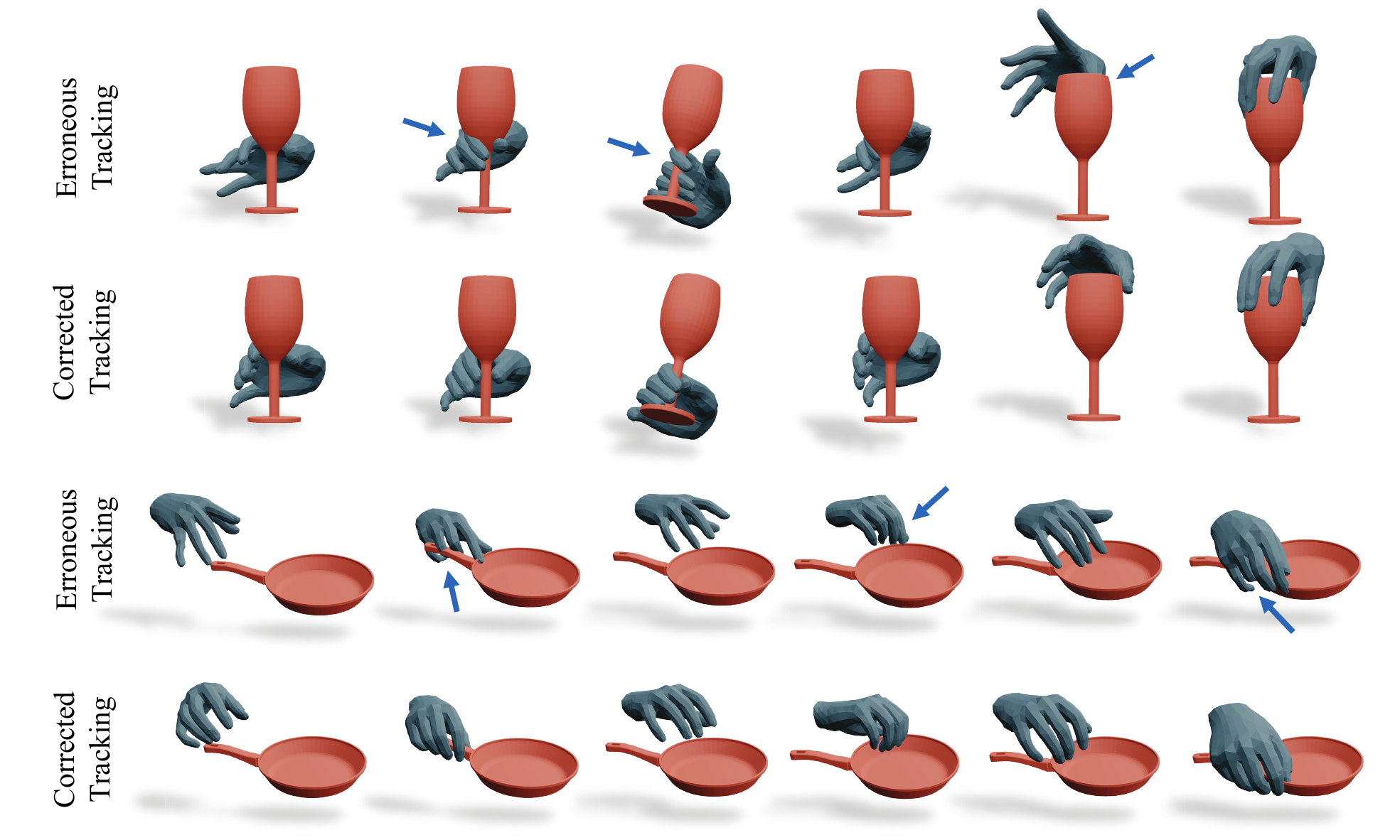}
    \caption{Qualitative results on two synthetic hand-object interaction sequences that suffer from inter-penetration and non-smooth hand motion. The results after TOCH refinement show correct contact and are much more visually plausible. Note that TOCH only applies minor changes in hand poses but the perceived realism is largely enhanced. Check the supplemental video for animated results.}
    \label{fig:grab}
\end{figure*}

\begin{table*}[t]
\small
\centering
\begin{tabular}{l|c|c|c|c|c}
\toprule
\shortstack[c]{GRAB-Type $\rightarrow$ \\ Noise $\rightarrow$} &  \shortstack[c]{GRAB-T \\ (0.01)} & \shortstack[c]{GRAB-T \\ (0.02)} & \shortstack[c]{GRAB-R \\ (0.3)} & \shortstack[c]{GRAB-R \\ (0.5)} & \shortstack[c]{GRAB-B \\ (0.01 \& 0.3)}  \\ 
\midrule
\vspace{0.1cm} MPJPE (mm) $\downarrow$  &  16.0 $\rightarrow$ \textbf{9.93} &  31.9 $\rightarrow$ \textbf{12.3} & \textbf{4.58} $\rightarrow$ 9.58  & \textbf{7.53} $\rightarrow$ 9.12 &  17.3 $\rightarrow$ \textbf{10.3}  \\ 
\vspace{0.1cm} MPVPE (mm) $\downarrow$  &  16.0 $\rightarrow$ \textbf{11.8} &  31.9 $\rightarrow$ \textbf{13.9} & \textbf{6.30} $\rightarrow$ 11.5  & \textbf{10.3} $\rightarrow$ 11.0 & 18.3 $\rightarrow$ \textbf{12.1} \\ 
\vspace{0.1cm} IV  ($\text{cm}^3$) $\downarrow$ &  2.48 $\rightarrow$ \textbf{1.79} & \textbf{2.40} $\rightarrow$ 2.50  &  1.88 $\rightarrow$ \textbf{1.52}& 1.78 $\rightarrow$ \textbf{1.35}&  2.20 $\rightarrow$ \textbf{1.78} \\ 
 C-IoU (\%) $\uparrow$  &  3.56 $\rightarrow$ \textbf{29.2} &  2.15 $\rightarrow$ \textbf{16.7}& 11.4 $\rightarrow$ \textbf{26.6} & 5.06 $\rightarrow$ \textbf{24.4} &  1.76 $\rightarrow$ \textbf{26.6} \\ 
\bottomrule
\end{tabular}
\caption{We quantitatively evaluate TOCH on multiple perturbed GRAB test sets with different types and magnitude of noise. The numbers inside the parentheses indicate standard deviation of the sampled Gaussian noise. Although pose accuracy is not always improved, TOCH significantly boosts interaction realism for all noise levels, which is demonstrated by the increase in contact IoU and reduction in hand-object inter-penetration.}
\label{tab:synthetic}

\end{table*}

\begin{table*}[t]
\centering
\begin{tabular}{l|ccc}
\toprule
\multirow{2}{*}{\textbf{Method}} & \multicolumn{3}{c}{\textbf{HO-3D}}  \\
&  MPJPE (mm) $\downarrow$  &MPVPE (mm) $\downarrow$ & IV ($\text{cm}^3$) $\downarrow$ \\
\midrule
Hasson \etal & 11.4& 11.4& 9.26 \\
RefineNet & 11.6& 11.5& 8.11\\
ContactOpt & 9.47& 9.45& 5.71\\
TOCH (ours) & \textbf{9.32}& \textbf{9.28}& \textbf{4.66}\\
\bottomrule
\end{tabular}
\caption{Quantitative evaluation on HO-3D compared to Hasson~\etal~\cite{hasson2020leveraging}, RefineNet~\cite{taheri2020grab} and ContactOpt~\cite{grady2021contactopt}. We follow the evaluation protocol of HO-3D and report hand joint and mesh errors after Procrustes alignment. TOCH outperforms all the baselines in terms of pose error and interaction quality.}
\label{tab:tracker}

\end{table*}

\subsection{Metrics}
\label{sec:exp_metrics}
\noindent
\textbf{Mean Per-Joint Position Error (MPJPE)}. We report the average Euclidean distance between refined and groundtruth 3D hand joints. Since pose annotation quality varies across datasets, this metric should be jointly assessed with other perceptual metrics.

\vspace{0.1cm}
\noindent
\textbf{Mean Per-Vertex Position Error (MPVPE)}. This metric represents the average Euclidean distance between refined and groundtruth 3D meshes. It assesses the reconstruction accuracy of both hand shape and pose.

\vspace{0.1cm}
\noindent
\textbf{Solid Intersection Volume (IV)}. We measure hand-object inter-penetration by voxelizing hand and object meshes and reporting the volume of voxels occupied by both. Solely considering this metric can be misleading since it does not account for the case where the object is not in effective contact with the hand.

\vspace{0.1cm}
\noindent
\textbf{Contact IoU (C-IoU)}. This metric assesses the Intersection-over-Union between the groundtruth contact map and the predicted contact map. The contact map is obtained from the binary hand-object correspondence by thresholding the correspondence distance within $\pm 2$ mm. We only report this metric on GRAB since the groundtruth annotations in HO-3D are not accurate enough~\cite{grady2021contactopt}.

\begin{figure*}[t]
    \centering
    \includegraphics[width=\linewidth]{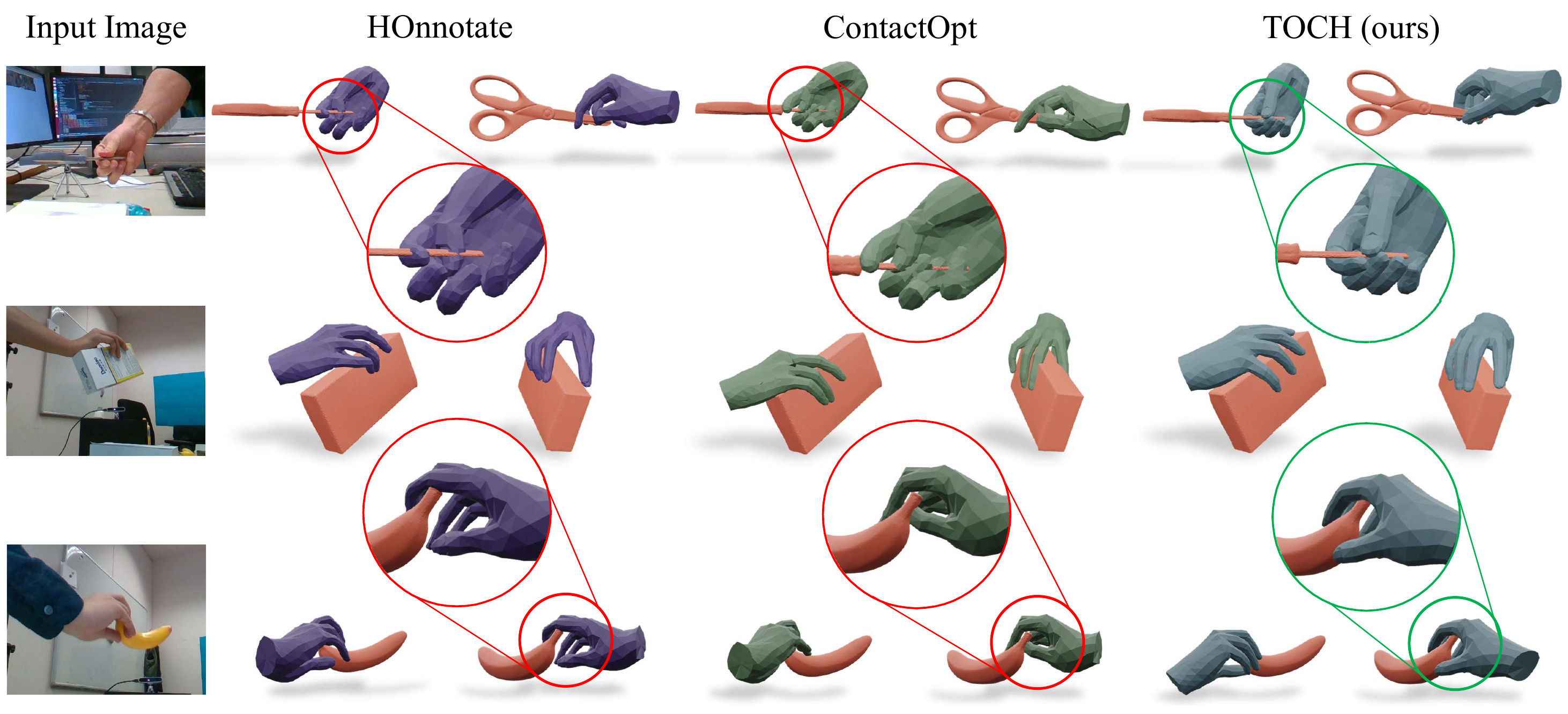}
    \caption{Qualitative comparison with HOnnotate and ContactOpt. Each sample reconstruction is visualized in two views, the image-aligned view and a side view. We can clearly see hand-object inter-penetrations for HOnnotate and ContactOpt, while our reconstructions are more visually realistic.
    }
    \label{fig:ho3d}
\end{figure*}

\subsection{Refining Synthetic Tracking Error}
\label{sec:exp_synthetic}
In order to use TOCH in real settings, it would be ideal to train the model on the predictions of existing hand trackers. However, this requires large amount of images/depth sequences paired with accurate hand and object annotations, which is currently not available. Moreover, targeting a specific tracker might lead to overfitting to tracker-specific errors, which is undesirable for generalization.

We observe that hand errors can be decomposed into inaccurate global translation and inaccurate joint rotations, and the inaccuracies produced by most state-of-the-art trackers are small. Therefore, we propose to synthesize tracking errors by manually perturbing the groundtruth hand poses of the GRAB dataset. To this end, we apply three different types of perturbation to GRAB: translation-dominant perturbation (abbreviated GRAB-T in the table) applies an additive noise to hand translation $\vec{t_H}$ only, pose-dominant perturbation (abbreviated GRAB-R) applies an additive noise to hand pose $\vec{\theta}$ only, and balanced perturbation (abbreviated GRAB-B) uses a combination of both. We only train on the last type while evaluate on all three. The quantitative results are shown in Table~\ref{tab:synthetic} and qualitative results are presented in Figure~\ref{fig:grab}. 

We can make the following observations. First, TOCH is most effective for correcting translation-dominant perturbations of the hand. For pose-dominant perturbations where the vertex and joint errors are already very small, the resulting hands after TOCH refinement exhibit larger errors. This is because TOCH aims to improve interaction quality of a tracking sequence, which can hardly be reflected by distance based metrics such as MPJPE and MPVPE. We argue that more important metrics for interaction are intersection volume and contact IoU. As an example, the perturbation of GRAB-R ($0.3$) only induces a tiny joint position error of $4.6$ mm, while it results in a significant $88.6\%$ drop in contact IoU. This validates our observation that any slight change in pose has a notable impact on physical plausibility of interaction. TOCH effectively reduces hand-object intersection as well as boosts the contact IoU even when the noise of testing data is higher than that of training data.

\subsection{Refining RGB(D)-based Hand Estimators}
\label{sec:exp_real}
To evaluate how well TOCH generalizes to real tracking errors, we test TOCH on state-of-the-art models for joint hand-object estimation from image or depth sequences. We first report comparisons with the RGB-based hand pose estimator~\cite{hasson2020leveraging}, and two grasp refinement methods RefineNet~\cite{taheri2020grab} and ContactOpt~\cite{grady2021contactopt} in Table~\ref{tab:tracker}. Hasson \etal~\cite{hasson2020leveraging} predict hand meshes from images, while RefineNet and ContactOpt have no knowledge about visual observations and directly refine hands based on 3D inputs. Groundtruth object meshes are assumed to be given for all the methods. TOCH achieves the best score for all three metrics on HO-3D. In particular, it reduces the mesh intersection volume, indicating an improved interaction quality.
We additionally evaluate TOCH on HOnnotate~\cite{hampali2020honnotate}, a state-of-the-art RGB-D tracker which annotates the groundtruth for HO-3D. Figure~\ref{fig:ho3d} shows some of its failure cases and how they are corrected by TOCH.

\begin{figure}[t]
    \centering
        \includegraphics[width=\textwidth]{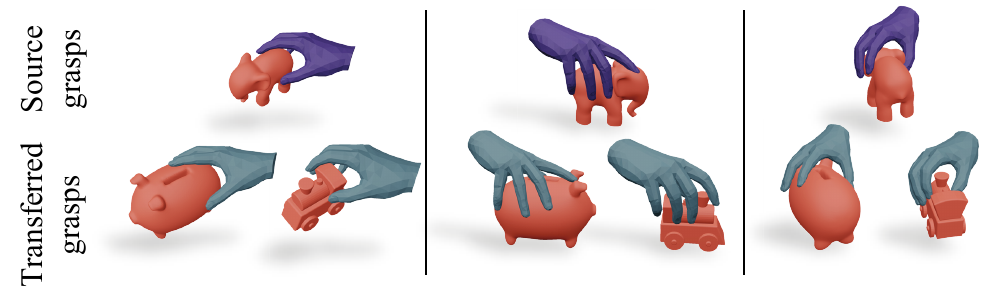}
        \caption{Transferring grasping poses across objects of different geometry. The top row shows three different source grasps which are subsequently transferred to two target objects in the bottom row. The hand poses are adjusted according to shape of target objects while preserving overall contact.}
        \label{fig:transfer}
\end{figure}

\begin{table*}[t]
\centering
\begin{tabular}{l|ccc}
\toprule
\textbf{Method} &  MPJPE (mm)  $\downarrow$ \quad\quad & IV ($\text{cm}^3$) $\downarrow$ \quad\quad & C-IoU (\%) $\uparrow$ \\
\midrule
Hand-centric baseline &  11.2&  2.03&18.9\\
TOCH (w/o corr.) &  12.2& 2.10& 18.6\\
TOCH (w/o GRU) & 10.8&  1.87& 20.4\\
TOCH (same obj.) & 11.7&  1.95& 23.1\\
TOCH (full model) & \textbf{10.3}& \textbf{1.78} & \textbf{26.6}\\
\bottomrule
\end{tabular}
\caption{Comparison with various baselines on GRAB-B (0.01 \& 0.3). We show that TOCH achieves the lowest hand joint error and intersection volume while recovers the highest percentage of contact regions among all the baselines.}
\label{tab:ablation}

\end{table*}

\subsection{Analysis and Ablation Studies}
\label{sec:exp_ablation}
\noindent
\textbf{Grasp transfer.} In order to demonstrate the wide-applicability of our learned features, we utilize the pre-trained TOCH auto-encoder for grasp transfer although it was not trained for this task.
The goal is to transfer grasping sequences from one object to another object while maintaining plausible contacts.
Specifically, given a source hand motion sequence and a source object, we extract the TOCH fields and encode them with our learned encoder network. We then simply decode using the target object -- we perform a point-wise concatenation of the latent vectors with point clouds of the target object, and reconstruct TOCH fields with the decoder. This way we can transfer the TOCH fields from the source object to the target object. Qualitative examples are shown in Figure~\ref{fig:transfer}.

\vspace{0.1cm}
\noindent
\textbf{Object-centric representation.} To show the importance of the object-centric representation, we train a baseline model which directly takes noisy hand joint sequences $\{\tilde{\vec{j}}^i\}_{i=1}^T$ as input and naively condition it on the object motion sequence $\{\vec{O}^i\}_{i=1}^T$. See Table~\ref{tab:ablation} for a quantitative comparison with TOCH. We can observe that although the hand-centric baseline makes small errors in joint positions, the resulting motion is less physically plausible, as reflected by its higher interpenetration and lower contact IoU. 

\vspace{0.1cm}
\noindent
\textbf{Semantic correspondence.} We argue that explicitly reasoning about dense correspondence plays a key role in modeling hand-object interaction. To show this, we train another baseline model in the same manner as in Section~\ref{sec:method}, except that we adopt a simpler representation $\vec{F}(\vec{H}, \vec{O}) = \{(c_i, d_i)\}_{i=1}^N$, where we keep the binary indicator and signed distance without specifying which hand point is in correspondence. The loss term (\ref{eq:loss_corr}) accordingly changes from mean squared error to Chamfer distance. We can see from Table~\ref{tab:ablation} that this baseline model gives the worst results in all three metrics.

\vspace{0.1cm}
\noindent
\textbf{Train and test on the same objects.} We test the scenario where objects in test sequences are also seen at training time. We split the dataset based on the action intent label instead of by objects. Specifically, we train on sequences labelled as 'use', 'pass' and 'lift`, and evaluate on the remaining. Results from Table.~\ref{tab:ablation} show that generalization to different objects works slightly better than generalizing to different actions. Note that the worse results are also partly attributed to the smaller training set under this new split.

\vspace{0.1cm}
\noindent
\textbf{Temporal modeling.} We verify the effect of temporal modeling by replacing the GRU layer with global feature aggregation. We concatenate the global average latent code with per-frame latent codes and feed the concatenated feature of each frame to a fully connected layer. As seen in Table.~\ref{tab:ablation}, temporal modeling with GRU largely improves interaction quality in terms of recovered contact.

\vspace{0.1cm}
\noindent
\textbf{Complexity and running time.} The main overhead incurred by TOCH field is in computing ray-triangle intersections, the complexity of which depends on the object geometry and the specific hand-object configuration. As an illustration, it takes around 2s per frame to compute the TOCH field on 2000 sampled object points for an object mesh with 48k vertices and 96k triangles on Intel Xeon CPU.
In hand-fitting stage, TOCH is significantly faster than ContactOpt since the hand-object distance can be minimized with mean squared error loss once correspondences are known.
Fitting TOCH to a sequence runs at approximately 1 fps on average while it takes ContactOpt over a minute to fit a single frame.

\section{Conclusion}
\label{sec:conclusion}
In this paper, we introduced TOCH, a spatio-temporal model of hand-object interactions.
Our method encodes the hand with TOCH fields, an effective novel object-centric correspondence representation which captures the spatio-temporal configurations of hand and object even before and after contact occurs. TOCH reasons about hand-object configurations beyond plain contacts, and is naturally invariant to rotation and translation.
Experiments demonstrate that TOCH outperforms previous methods on the task of 3D hand-object refinement.
In future work, we plan to extend TOCH to model more general human-scene interactions.

\label{sec:ack}
\vspace{0.3cm}
\noindent
\textbf{Acknowledgements} 
This work is supported by the German Federal Ministry of Education and Research (BMBF): Tübingen AI Center, FKZ: 01IS18039A. This work is funded by the Deutsche Forschungsgemeinschaft (DFG, German Research Foundation) - 409792180 (Emmy Noether Programme, project: Real Virtual Humans). Gerard Pons-Moll is a member of the Machine Learning Cluster of Excellence, EXC number 2064/1 – Project number 390727645. The project was made possible by funding from the Carl Zeiss Foundation.

\clearpage
%
%
\bibliographystyle{splncs04}
\bibliography{egbib}
\end{document}